\documentclass{article}

\usepackage{arxiv}

\usepackage[utf8]{inputenc} 
\usepackage[T1]{fontenc}    
\usepackage{hyperref}       
\usepackage{url}            
\usepackage{booktabs}       
\usepackage{amsfonts}       
\usepackage{nicefrac}       
\usepackage{microtype}      
\usepackage{lipsum}
\usepackage{microtype}
\usepackage{multicol}
\usepackage{graphicx}
\usepackage{xspace}
\usepackage{multirow}
\usepackage{xcolor}
\newcommand{\datasetName}{TechTrack\xspace}
\newcommand{\propara}{ProPara\xspace}
\newcommand{\sbcomment}[1]{}

\newcommand{\eat}[1]{}
\usepackage{paralist}
\usepackage{booktabs}

\title{Tracking Entities in Technical Procedures - \\A New Dataset and Baselines}

\author{
 Saransh Goyal, Pratyush Pandey, Garima Gaur, Subhalingam D, Srikanta Bedathur, Maya Ramanath\\
 Indian Institute of Technology, Delhi \\
 New Delhi, India 110016\\
}
\date{ } 
\begin{document}
\maketitle

\begin{abstract}
We introduce \datasetName, a new dataset for tracking entities in technical procedures. The dataset, prepared by annotating open domain articles from WikiHow, consists of 1351 procedures, e.g., ``How to connect a printer'', identifies more than 1200 unique entities with an average of 4.7 entities per procedure. We evaluate the performance of state-of-the-art models on the entity-tracking task and find that they are well below the human annotation performance. We describe how \datasetName can be used to take forward the research on understanding procedures from temporal texts.
\end{abstract}


\section{Introduction}
We present a new dataset, \datasetName, to advance research in the understanding of \emph{procedural text} -- i.e., text that describes a sequence of actions geared towards achieving an end goal. We focus on procedures from \emph{technical} ``How-to''s. Such text is typically seen in FAQs and manuals, which consist of step-by-step answers to questions such as ``How to print a test page on a printer?'' or ``How to troubleshoot a network connection?''. The answers consist of step-by-step instructions for completing the task or troubleshooting the problem in the question.

\begin{figure*}
    \centering
        \includegraphics[width=\textwidth]{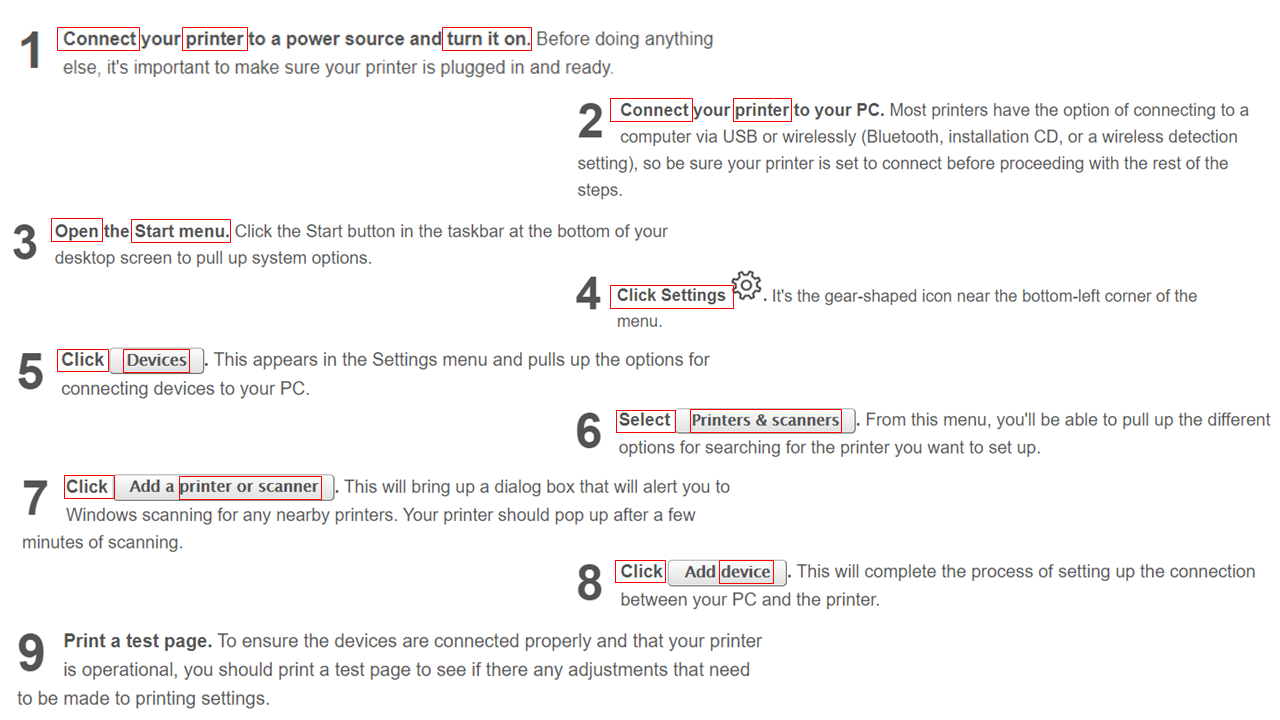}
    \caption{WikiHow: How to setup a printer. The highlighted portions of the steps describe entities and actions which result in changing states of the entities. For example, in step 1, the entity \emph{Printer} changes the value of its attribute \emph{isPoweredOn} from \emph{false} to \emph{true}.}
    \label{fig:wikihow}
\end{figure*}

This kind of procedural text has specific entities of interest (for example, \emph{printer}, \emph{printer driver}, \emph{ethernet card}, etc.), and these entities have attributes whose values change over the course of actions described in the text. An example procedure from the WikiHow website is shown in Figure \ref{fig:wikihow}. Several interesting entities are highlighted, along with actions that change an attribute value of those entities. The entities themselves could be hardware entities, such as \emph{Printer} or software entities such as the \emph{Settings} button. The \emph{entity tracking problem} is the problem of identifying the changing values of the attributes of interesting entities in a procedure. This is a challenging problem and is one of the first steps towards understanding procedures.\sbcomment{do we have a reference to say how this is "first step toward understanding of procedural text" ?}

Previous work in this area has proposed datasets such as bAbI \cite{babi} (about narratives and stories) and SCoNE \cite{scone}. However, due to their dependency on the synthetic text generators, they fail to capture the intricacies of real world procedures. Two recently introduced real procedural datasets for the entity tracking problem include \propara \cite{Dalvi_2018}, and the NPN-Cooking dataset \cite{bosselut2017simulating}. However, the former is concerned with scientific processes rather than step-by-step procedures, while the latter is an action-centric dataset with a data model that is not applicable in wider settings. The cooking dataset associates the actions with the properties of entities they might affect. For example, action WASH can only affect the \emph{cleanliness} property of entities. However, in \datasetName simply associating a set of entity properties with an action is not possible. For instance, in instructions, ``Connect the printer to computer" and  ``Connect phone to WiFi", action CONNECT would impact the different properties of entities.
\subsection{Our Contributions}
In this paper, we describe \datasetName dataset derived from 1351 procedures from one technical category of Wikihow. Each of the multi-step procedures are annotated with entities, their properties, and the value of these properties. On this dataset, we evaluate the performance of entity-tracking task using a BERT-based baseline we developed and the recently proposed entity tracking model, ProLocal \cite{Dalvi_2018}. We find that there is a significant gap to cover (Table \ref{tab:TrueF1All} and \ref{tab:combined}).
We release \datasetName  to the community as a challenging realistic procedural text dataset. The TechTrack dataset will be made available freely at~\url{http://github.com/<anonymized>}. 

\sbcomment{Now that I am reading this afresh, I feel that the intro does not tell me (a) what can be gained by solving this problem? (b) is the characteristic of technical procedures seen elsewhere as well? Basically, why should any one care about using this specific benchmark dataset?}

\section{Description of \datasetName}

\begin{table*}[ht]
\centering
{
  \begin{tabular}{llr}
    \toprule
\textbf{Category Name}&\textbf{Examples}&\textbf{Properties}\\ \midrule
    Hardware-Devices & Printer, Scanner & isPowered, isConnected, \eat{isSetup,} isUsed\\
    Software-Device Drivers & Printer Driver & isInstalled, \eat{isRelatedDvcCnnctd,} isSettingsChanged\\
    Software-OS Related & File Explorer, Settings Menu & isOpened, isSettingsChanged\\
    Software-Other & Chrome, Adobe PDF Reader & isInstalled, isOpened, isSettingsChanged\\
\bottomrule
\end{tabular}
}
\caption{Entity categories and the annotated properties for those categories derived from WikiHow in the \datasetName dataset.}
  \label{tab:props}
\end{table*}

\begin{table*}[ht]
\centering
\resizebox{\textwidth}{!}
{\begin{tabular}{l|lll|c|c|c|ll|}
\cline{2-9}
\multicolumn{1}{c|}{\textbf{}} & \multicolumn{3}{c|}{\textbf{Printer}} & \textbf{StartMenu} & \textbf{Settings} & \textbf{Devices} & \multicolumn{2}{c|}{\textbf{Printers \& Scanners}} \\ \hline
\multicolumn{1}{|c|}{\textbf{Setup a Printer}} & \multicolumn{1}{c}{\textbf{isPow}} & \multicolumn{1}{c}{\textbf{isConn}} & \multicolumn{1}{c|}{\textbf{isSetup}} & \textbf{isOpened} & \textbf{isOpened} & \textbf{isOpened} & \multicolumn{1}{c}{\textbf{isOpened}} & \multicolumn{1}{c|}{\textbf{isSettings}} \\ \hline
\multicolumn{1}{|l|}{0 - Setting Up a Printer on Windows} & False & False & False & False & False & False & False & False \\
\multicolumn{1}{|l|}{\begin{tabular}[c]{@{}l@{}}1- Connect your printer to a power \\ source and turn it on.\end{tabular}} & True & False & False & False & False & False & False & False \\
\multicolumn{1}{|l|}{2 - Connect your printer to your PC.} & True & True & False & False & False & False & False & False \\
\multicolumn{1}{|l|}{3 - Open the Start menu.} & True & True & False & True & False & False & False & False \\
\multicolumn{1}{|l|}{4 - Click Settings .} & True & True & False & True & True & False & False & False \\
\multicolumn{1}{|l|}{5 - Click Devices.} & True & True & False & True & True & True & True & False \\
\multicolumn{1}{|l|}{6 - Select Printers \& scanners .} & True & True & False & True & True & True & True & True \\
\multicolumn{1}{|l|}{7 - Click Add a printer or scanner .} & True & True & False & True & True & True & True & False \\
\multicolumn{1}{|l|}{8 - Click Add device.} & True & True & True & True & True & True & True & False \\
\multicolumn{1}{|l|}{9 - Print a test page.} & True & True & False & True & True & True & True & False \\ \hline
\end{tabular}
}
\caption{Annotations for the procedure ``How to setup a printer''. For ease of understanding, only the heading of each step is shown. The first row lists a few entities of interest. For each entity, the property values are tabulated.}
  \label{tab:annotations}
\end{table*}
\normalsize

In preparing \datasetName, we focused on the task of \textbf{entity tracking}, defined as follows: Given a procedure, denoted by a sequence of steps $(s_i)$, a set of entities $\{e_i\}$ and a pre-defined set of properties $\{a_{ij}\}$, where $a_{ij}$ corresponds to the $j^{th}$ property of entity $e_i$, identify the value of each property at each step $s_i$ in the procedure.

We prepared \datasetName by extracting procedures from the `Computers and Electronics' category of the WikiHow website which consists of multiple topics such as ``Windows'', ``Printers'', ``Linux'', ``Monitors'', etc. Table \ref{tab:props} shows the categories of entities and properties that are tracked for entities of each category. The entity, properties and property value annotations were all crowdsourced. The annotators were informed that every entity should be categorized into one of 4 categories, what properties to track for each category, and what values were valid. 

Table \ref{tab:annotations} shows an example procedure and the corresponding annotations in tabular form. We also note two kinds of properties: \emph{state} and \emph{event}. In the former, unless some action is taken, the property value remains the same. For example, if a printer is turned ’on’ in a particular step, it remains ’on’ for the rest of the procedure until another action is taken. For event properties, the values from a step are not carried forward to the next steps. For example, if the printer is used to print something in the current step, the \emph{isUsed} property changes only in this step and not in future steps.

The final dataset contains annotations for 1351 procedures, identifies over 1200 unique entities, with an average of 4.7 entities per procedure. Further, each procedure contains an average of 9 steps, and each step has an average of 51 words (split into multiple sentences).

In each step of the procedure, values of one or more properties of one more entity may change. Our model should be able to correctly output this change. One of the key differences from the ProPara dataset \cite{Dalvi_2018} is that the set of properties and their possible values are fixed and known beforehand. This ensures that the model is tested solely on its ability to detect the changes in property values.

This annotated dataset can be used to evaluate tasks other than entity tracking as well. For example, a common problem in automated technical support systems is to suggest to the user ``what to do next'' after performing a sequence of steps (or reaching a specific state over all relevant entities and their properties). 
Our dataset can naturally be used in this setting.

\section{Experiments}

\paragraph*{Baselines}
We used our dataset to perform experiments with the following: a recently proposed model, ProLocal \cite{Dalvi_2018}, and a model we adapted based on BERT \cite{devlin2018BERT}. Neither model is directly applicable to our setting, and so we adapted them as follows.

\begin{description}

\item[ProLocal:] The ProLocal model predicts \emph{text spans} corresponding to property values, rather than property values themselves. Further, the model only classifies a single property. Since we have a pre-defined set of property values, we only make use of ProLocal's state change classifier and train separate versions of this model for each property.

\item[BERT:] The BERT-based model only accepts natural language queries. Therefore, we cast our problem into asking template-based questions such as ``Is the printer powered on?'' to determine the value of property \emph{isPoweredOn} for the entity \emph{printer}. For a fair comparison, we build separate versions of the model for each property.
\end{description}
\subsection{Evaluation Metrics}
We assumed that the default value of all state-based properties is ``None'' of ``No change'', while event-based properties are intially ``False''. Therefore, we need to measure the performance of the models in correctly predicting a change in state-based properties and a ``Going-to-true'' state (that is, at the end of the step, the value of the property is ``True''). The opposite can also occur (that is, a change from ``True'' to ``False''), but since this occurs in less than $1\%$ of the state changes, we ignore them.

\subsection{Results}
\begin{description}
\item[Setup 1: Training each property Individually] In our first setup, we trained the models for each property individually. Table \ref{tab:TrueF1All} shows the list of properties and the amount of training and test data for each row.  Clearly, the BERT-based model outperforms ProLocal. An interesting observation, however, is that the BERT-based model performs poorly for the event property type \emph{isSettingsChanged} compared to ProLocal despite having quite a lot of training data.

This approach suffers from the size bias of the dataset: some properties such as \emph{isUsed} have very limited training data, which causes the prediction accuracies to be quite poor. To combat this problem, we train the model jointly on all properties next.

\begin{table*}[ht]
\centering

{\begin{tabular}{llllrrrrrr}
\toprule
\multirow{2}{0.5in}{\textbf{Property Name}}&\multirow{2}{0.5in}{\textbf{Property Type}}&\multirow{2}{0.5in}{\textbf{Train Rows}} & \multirow{2}{0.5in}{\textbf{Test Rows}} &   \multicolumn{3}{c}{\textbf{BERT}} & \multicolumn{3}{c}{\textbf{ProLocal}}\\
\cmidrule(lr){5-7}\cmidrule(lr){8-10}
&&&& \textbf{Prec} & \textbf{Rec}&\textbf{F1}&\textbf{Prec}&\textbf{Rec}&\textbf{F1} \\
\midrule

{isOpened}                 & State & 36248  & 5014 & 0.76 & 0.89 & {\bf 0.82} & 0.62 & 0.76 & 0.68 \\
{isSettingsChanged}        & Event & 10846 & 1635 & 0.41 & 0.55 & 0.47 & 0.55 & 0.72 & {\bf 0.62} \\
{isPowered}                & State & 1488 & 158 & 0.64 & 0.47 &  0.55 & 0.75 & 0.46 & {\bf 0.57} \\
{isInstalled}              & State & 2334 & 310 & 0.65 & 0.69 & {\bf 0.67} & 0.46 & 0.75 & 0.57 \\
{isConnected}              & State & 709  & 62  & 0.33 & 0.36 & {\bf 0.35} & 0.2  & 0.14 & 0.17 \\
{isUsed}                   & Event & 606  & 94 & 0.29 & 0.2  & {\bf 0.24} & 0    & 0    & 0    \\
\bottomrule
\end{tabular}}
\caption{\label{tab:TrueF1All}
Precision, Recall and F1 for each property.
}
\end{table*}
\normalsize
%
%
\item[Setup 2: Training the Combined Model] Along with individual models for each property, we also train a combined BERT-based classifier to predict property values for each property type. Here, we train a single model on multiple properties, with each property trained on respective training rows. As seen in Table \ref{tab:combined}, the model results in considerable boost in performance of properties with lesser data, mainly \emph{isUsed} and \emph{isSetup}. This improvement is partly caused by the flow of information from other properties and partly due to better training of the language model of BERT.


\begin{table}[]
    \centering
    \begin{tabular}{lrrrr}
    \toprule
    \textbf{Property} & \textbf{Prec} & \textbf{Rec} & \textbf{F1} & \textbf{Change} \\
    \midrule
    isOpened             & 0.75 & 0.88 & 0.81 & - 0.01 \\
    isSettingsChanged    & 0.7  & 0.54 & 0.61 & 0.14   \\
isPowered                & 0.44 & 0.69 & 0.54 & - 0.01 \\
isInstalled              & 0.61 & 0.78 & 0.68 & 0.01  \\
isConnected              & 0.39 & 0.69 & 0.5  & 0.15  \\
isUsed                   & 0.75 & 0.71 & 0.73 & 0.49  \\
    \bottomrule
    \end{tabular}
\caption{Performance of combined BERT-based model on all properties, compared with individual BERT-based models trained over each property.}
\label{tab:combined}
\end{table}


\item[Setup 3: Performance for different topics] Our final experiment was to see if the accuracy of the models for a given property changed if the procedure was for different topics. Recall that TechTrack was constructed from WikiHow category 'Computers and Electronics' which consists of multiple topics. Table \ref{tab:folders} reports the accuracy for the \emph{isOpened} property on these topics. The $F1$ values varies over the different topics, from $0.66$ for ``Linux'' to $0.92$ for ``OS''. Further investigation is required to identify the reasons for this.

\end{description}
\begin{table}[ht]
\centering
\begin{tabular}{lllll}
\toprule 
\textbf{Topics} & \textbf{TestRow} & \textbf{Prec} & \textbf{Rec} & \textbf{F1} \\ \midrule
Ubuntu               & 404                & 0.8           & 1            & 0.89        \\
Webcams              & 210                & 0.81          & 1            & 0.9         \\
Windows              & 1662               & 0.82          & 0.93         & 0.87        \\
Linux                & 245                & 0.57          & 0.77         & 0.66        \\
Mac                  & 959                & 0.72          & 0.87         & 0.79        \\
OS                   & 219                & 0.85          & 1            & 0.92        \\
Printers             & 168                & 0.8           & 0.95         & 0.87        \\
Monitors             & 245                & 0.875         & 0.875        & 0.875       \\
OSX                  & 462                & 0.86          & 0.98         & 0.92        \\
Windows10            & 360                & 0.77          & 0.96         & 0.85     \\  \bottomrule
\end{tabular}
  \caption{Performance of the BERT-based model on the \emph{isOpened} property on various topics from the dataset}
  \label{tab:folders}
\end{table}

\section{Related Work}

Procedural text, due to its inherent difference from the factual text, has lead to new datasets and models specially designed for the comprehension task. bAbI \cite{babi} (about narratives and stories) and SCoNE \cite{scone} were some of the early procedural text datasets. 

Two recently introduced realistic procedural datasets are \propara \cite{Dalvi_2018} and NPN-Cooking dataset \cite{bosselut2017simulating}. \propara contains  $488$ handwritten paragraphs about scientific processes. The NPN-Cooking dataset has around $65K$ recipes with involved ingredients as entities. \propara dataset is supplied with two models, ProLocal and ProGlobal, that are trained to identify the location and state of relevant entities at the sentence level and global level, respectively. Similar to factual question answering systems, these models report a text span as the location of the entities. A few significant efforts towards solving this location tracking task includes -- DynaPro \cite{amini2020procedural} an attention-based model that jointly predicts the property and the transition of the entities, KG-MRC \cite{das2018building} that captures the evolution of entities using a dynamic knowledge graph and leverages machine reading comprehension (MRC) models for locating the relevant text spans, Pro-Struct \cite{tandon2018reasoning} model augmented with commonsense constraints to enhance the quality of predictions, \cite{bosselut2017simulating} an attention-based model that tracks changes induced by actions in the procedure.

Our dataset \datasetName, unlike any of these datasets, is focused on technical procedures, such as those found in How-to manuals, has a pre-defined set of entity categories, and is property-centric.

\section{Conclusions}
We were motivated by the limited availability of datasets for procedural text understanding in real-life scenarios. \datasetName is created to be a realistic dataset that contains not only a diverse set of entities but also multiple properties whose value changes have to be tracked over the course of a procedure. It is richer than ProPara, the closest comparable dataset. Our evaluation of state-of-the-art models for entity tracking on \datasetName shows a significant performance gap that needs to be filled.



\end{document}